\documentclass{article}
\usepackage{styleij}

\usepackage{times}

\usepackage{soul}
\usepackage{url}
\usepackage[hidelinks]{hyperref}
\usepackage[utf8]{inputenc}
\usepackage[small]{caption}
\usepackage{graphicx}
\usepackage{amsmath}
\usepackage{booktabs}
\usepackage{booktabs}
\usepackage{subfigure}
\usepackage{threeparttable}
\usepackage{multirow}
\usepackage{color}
\usepackage{subfigure}
\usepackage{threeparttable}
\usepackage{multirow}
\usepackage{color}
\newcommand{\tabincell}[2]{\begin{tabular}{@{}#1@{}}#2\end{tabular}}

\title{Direction-aware Feature-level Frequency Decomposition for Single Image Deraining}

\author{
Sen Deng$^{1,}$\thanks{Joint first authors}\and
Yidan Feng$^{1,}$\footnotemark[1]\and
Mingqiang Wei$^{1,}$\footnote{Corresponding authors: mingqiang.wei/hrxie2@gmail.com}\and
Haoran Xie$^{2,}$\footnotemark[2]\and
Yiping Chen$^3$\and
Jonathan Li$^3$\and
Xiao-Ping Zhang$^4$\And
Jing Qin$^5$\\
\affiliations
$^1$Nanjing University of Aeronautics and Astronautics, Nanjing, China\\ 
$^2$Lingnan University, Hong Kong, China\\
$^3$Xiamen Univeristy, Xiamen, China\\
$^4$Ryerson University, Toronto, Canada\\
$^5$Hong Kong Polytechnic University, Hong Kong, China\\
}

\begin{document}

\maketitle

\begin{abstract}
We present a novel direction-aware feature-level frequency decomposition network for
single image deraining.
Compared with existing solutions, the proposed network has three compelling characteristics.
First, unlike previous algorithms, we propose to perform frequency decomposition at feature-level instead of image-level, allowing both low-frequency maps containing structures and high-frequency maps containing details to be continuously refined during the training procedure.  
Second, we further establish communication channels between low-frequency maps and high-frequency maps to interactively capture structures from high-frequency maps and add them back to low-frequency maps and, simultaneously, extract details from low-frequency maps and send them back to high-frequency maps, thereby removing rain streaks while preserving more delicate features in the input image.  
Third, different from existing algorithms using convolutional filters consistent in all directions, we propose a direction-aware filter to capture the direction of rain streaks in order to more effectively and thoroughly purge the input images of rain streaks.
We extensively evaluate the proposed approach in three representative datasets and experimental results corroborate our approach consistently outperforms state-of-the-art deraining algorithms.
\end{abstract}

\section{Introduction}
Images captured in rainy weather suffer from severe visibility degradation, which may impose great negative effects on many computer vision tasks, including object detection and tracking, autonomous driving, and semantic segmentation.  
%
In this regard, image deraining is an essential prerequisite for many vision applications, seeking to recover the clean image from its complex entanglement with rain streaks.
This problem is, however, very challenging and ill-posed, as the underlying background is totally unknown.

Many efforts have been dedicated to addressing the problem.
Early investigations are mainly based on various image priors.
One of the image priors closely related to deraining is that the main structures of an image are usually of low frequency while the details, such as rain streaks, are often of high frequency~\cite{perona1990scale}.
Naturally, the pioneering work of image deraining first adopts bilateral filtering to decompose the images into low frequency maps and high frequency maps, and then deal with rain streaks in the high frequency maps using dictionary learning~\cite{p_fu2011single}.
Later, more approaches have been proposed based on this prior~\cite{p_wang2017hierarchical}, as well as other image priors, such as Gaussian mixture model~\cite{GMM} and low-rank representations~\cite{low_rank_respresentation_1}. 
However, these hand-crafted image priors are incapable of disentangling structures, particularly exquisite ones, from rain streaks, thereby reconstructing unsatisfactory clean images.   
%

\begin{figure*}[!t] \centering
	\includegraphics[height=0.3\linewidth, width=0.9\linewidth]{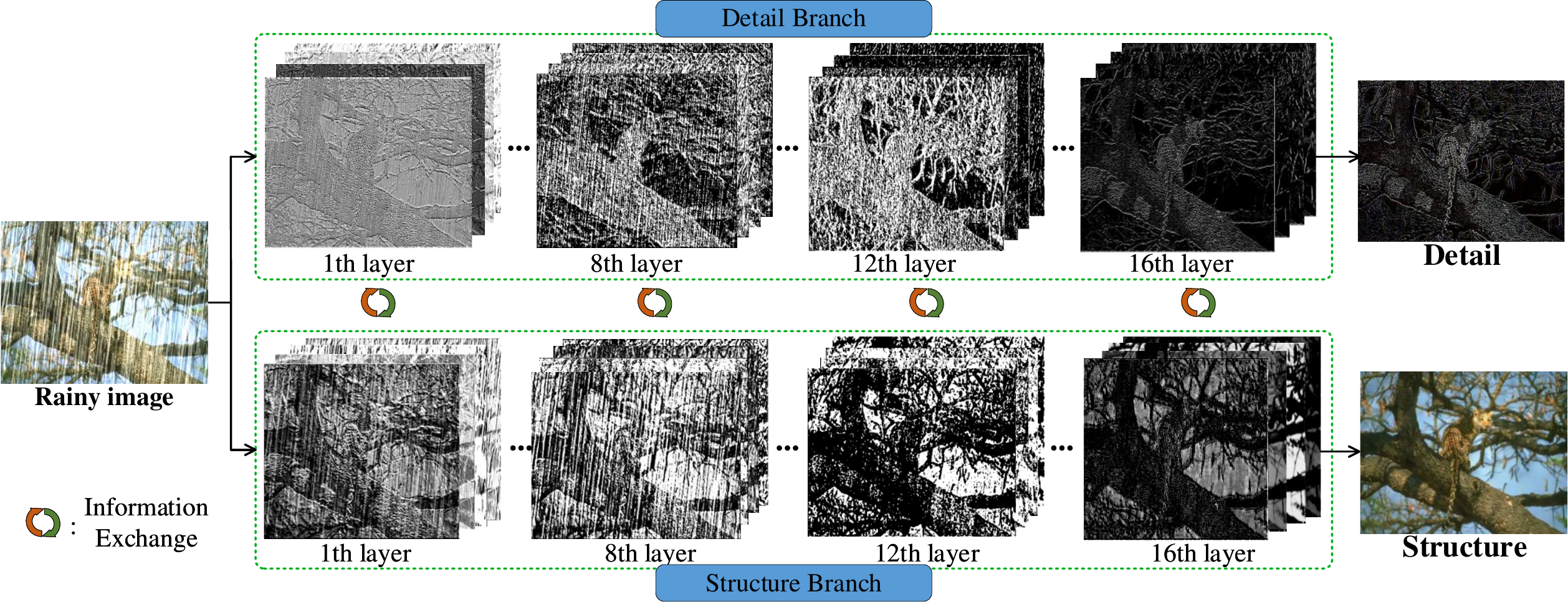}
	\caption{Visualization of feature maps in progressive frequency decomposition. The first row refers to the extraction of high-frequency details from the rainy image. As the layer goes deeper, the feature maps emphasize more and more on the foreground and are less entangled with rain streaks. While in the low-frequency structure branch, the network gradually obtains clean features highlighted on the background.
	}
	\label{fig:introduction}
\end{figure*}\

Recently, the performance of image deraining is boosted by deep convolutional neural networks (CNNs), which aim at capturing a variety of image characteristics by learning a complex model from massive data.
The first deep learning-based deraining framework is proposed by~\cite{DDN}.
In this work, after a frequency-based decomposition operation, a three-layer CNN is directly adopted to extract rain streaks from the high-frequency maps. 
Later, more CNN models have been proposed by introducing either extra network modules~\cite{PRENET} or task-specific auxiliary information~\cite{DAF-Net,d_zhang2018density} to guide the learning process, attempting to capture more powerful features to distinguish image structures and rain streaks.
However, these models still have several shortcomings.
First, these approaches perform frequency decomposition only at the image level, making the rain streaks mistakenly assigned to low-frequency maps difficult to be effectively removed and, simultaneously, the delicate structures assigned to high-frequency maps difficult to be recovered in the clean images.  
Second, there lack of interactive mechanisms between the low-frequency maps and high-frequency maps in the training procedure.
Third, traditional convolutional filters are often consistent in all directions while the rain streaks in an image usually head in one direction, i.e. the wind direction; this property is ignored by most of current solutions.

In order to comprehensively address these shortcomings, in the paper, we propose a novel network with dual branches for single image deraining.
Compared with existing solutions, the proposed network has three compelling characteristics.
First, unlike previous algorithms, we propose to perform frequency decomposition at feature-level instead of image-level.
By this way, the proposed network is able to generate low-frequency maps and high-frequency maps from feature maps at different layers, and hence allows these maps to be continuously refined during the training procedure (see Fig. \ref{fig:introduction}).
Second, we further establish communication channels between the dual branches, promoting the information propagation between low-frequency maps and high-frequency maps during the training procedure. 
Such a mechanism is not only helpful for separating more rain streaks from low-frequency maps to facilitate deraining but also useful for extracting more delicate features from high-frequency maps and add them back to low-frequency maps, enhancing clean image reconstruction.     
Third, most existing methods employ convolutional filters harmonious in all directions but ignore the fact that the rain streaks always head to the wind direction in an image, and hence are sub-optimal for image deraining. 
In order to take full advantage of this phenomenon, we propose a novel cross-median filter to capture the direction of rain streaks, aiming at producing more representative features to thoroughly purge the input image of rain streaks.
We extensively evaluate the proposed network on three famous image deraining datasets.
Experimental results demonstrate the effectiveness of the proposed network, consistently outperforming state-of-the-art approaches in most metrics.
Our contributions can be summarized as:
\begin{itemize}
	\item We propose a novel network with dual branches for single image deraining, conducting frequency decomposition at feature level instead of image level so as to gradually and iteratively refine both low frequency maps and high frequency maps during training.
	\item We propose a new mechanism to promote interactions between low frequency maps and high frequency maps, facilitating both rain streak removal and fine features recovery; we further propose a novel direction-aware filter to more efficiently and effectively capture rain streaks in training. 
	\item We set state-of-the-art performance of single image deraining on three famous datasets.
\end{itemize}

\begin{figure*}[!t] \centering
	\includegraphics[height=0.26\linewidth, width=1\linewidth]{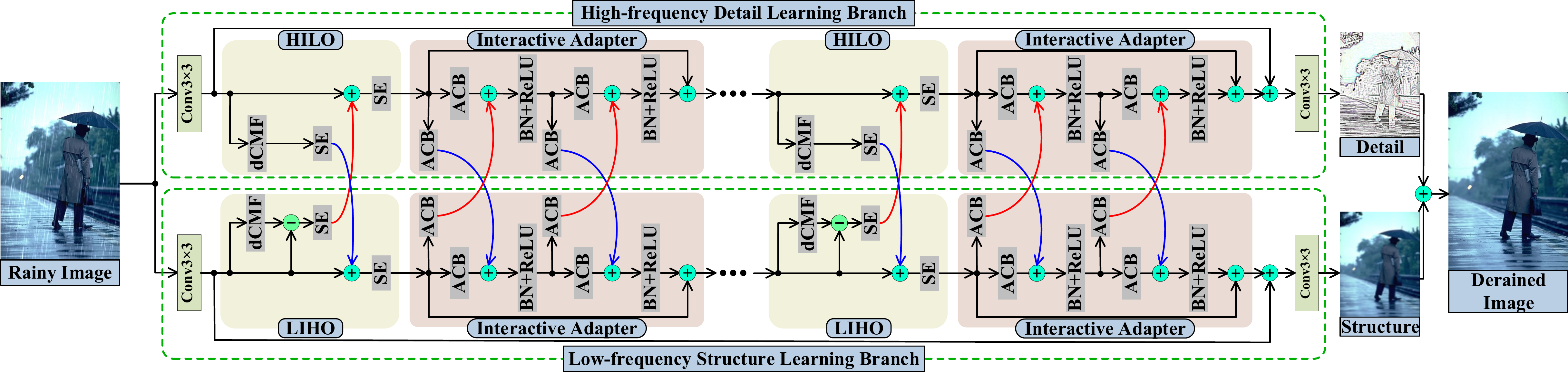}
	\caption{Our proposed network learns decomposed labels via two parallel yet interactive branches, where a detail learning branch keeps peeling off low-frequency components while reusing high-frequency features stripped from a structure learning branch, and vice versa. The two branches share similar structures. For detail learning branch, features are first fed into a High In Low Out (HILO) module, which extracts the low-frequency (blue arrow) using the proposed direction-aware Cross-Median Filter (dCMF) and accepts the high-frequency (red arrow) from a corresponding Low In High Out (LIHO) module, and then an Interactive Adapter based on Asymmetric Conv Block (ACB) is used for feature learning and adapting. This procedure is conducted iteratively for robust and effective learning on both clean details and structures.}
	\label{fig:network_framework}
\end{figure*}

\section{Related Work}
\subsection{Conventional Methods}
Mainstream in conventional methods models image deraining as an image decomposition problem, where a rainy image is decomposed into a clean background layer and a rain streak layer \cite{p_fu2011single}. This strategy is followed by \cite{p_wang2017hierarchical}, which introduce more prior knowledge, such as depth of field  and color variance, for better extracting the rain streaks from the detail layer. In addition, many other image priors are exploited for image deraining. \cite{p_chen2013generalized} first adopts low rank representation to describe the non-local similarity in different rain patches, which is further explored by \cite{low_rank_respresentation_2}. \cite{DSC} considers the difference in rain streak layer and background layer, based on which they propose a novel discriminative sparse coding method. \cite{GMM} exploits Gaussian mixture models for rain removal, which is learned on small patches that can accommodate a variety of background appearances and rain streak appearances. \cite{p_gu2017joint} combines analysis sparse representation and synthesis sparse representation to better separate rain streaks and image textures.

\subsection{Deep Learning-based Methods}
Using frequency domain decomposition and residual connections, \cite{DDN} first employs a three-layer CNN to extract rain streaks form the detail layer. Thereafter, advancing network modules are introduced, such as residual block \cite{d_deng2019drd}, dilated convolution \cite{d_deng2019drd} and recursive block \cite{PRENET}. Among them, \cite{acm_yu2019gradual} and adopts a coarse-to-fine strategy by adding supervision on different learning stages. Due to the complexity of rain streaks and their composition with the background, several methods are proposed to separate the task using dual-path networks \cite{d_deng2019drd,DUAL}, or adopt a multi-stage strategy using recurrent neural networks to progressively recover the clean image \cite{RESCAN,PRENET}. \cite{DUAL} proposes to recover low frequency image structures and high frequency image details separately using two parallel network branches. \cite{d_deng2019drd} takes advantage of another network branch to find back lost details. GAN is also exploited by \cite{d_pu2018cyclegan,RAIN800} to refine the deraining results for more visual appealing effects. Besides, \cite{DAF-Net}
builds a dataset to describe heavy rainy scenes using depth images to associate rain streaks and rainy haze. \cite{SPA-Net} proposes a real rain dataset using video-based deraining results and adopt a directional IRNN to learn spatial attention for guiding the network. \cite{d_li2019bench} presents a comprehensive benchmark named MPID for evaluation of various deraining methods. \cite{zhu2019singe} first utilizes CycleGAN for single image deraining. For removing different scales of rain streaks, \cite{yang2020towards} designs a fractal band learning network trained with self-supervision for scale-robust rain streak removal.

\section{Method}
In this section, we introduce our proposed method built on frequency decomposition. Denote the rain-free label image as $\mathbf{I}_{gt}$, it is first decomposed into a low-frequency structure map $\tilde{\mathbf{S}}$ and a high-frequency detail map $\tilde{\mathbf{D}}$. Our goal is to accurately predict both $\tilde{\mathbf{S}}$ and $\tilde{\mathbf{D}}$ using a single rainy input $\mathbf{I}$, so as to recover a high-fidelity derained image with abundant details and minimum distortions. Our solution resorts to feature-level frequency decomposition along and across a parallel network architecture. The extraction of a particular frequency is learned along each branch towards the decomposed label, while components of the other frequency are continuously delivered across the branch. We first interpret the multiple labels and their loss functions, and then go into details of the direction-aware Cross-Median Filter (dCMF) and the interactive adapter to explain how we isolate the different frequency to enhance communication between branches.

\subsection {Label Decomposition and Loss Functions}
Label decoupling in low-level vision is a strategy that models the final task as a composition of several easier sub-tasks \cite{wei2020label,DUAL}, which we argue is particularly effective in image deraining due to the complex entanglement of rain streaks and image contents \cite{d_deng2019drd}. In our case, the label image $\mathbf{I}_{gt}$ is decomposed into low- and high- frequencies using a low-pass image filter, where the high frequency part $\tilde{\mathbf{D}}$ contains abundant image details and the low frequency part $\tilde{\mathbf{S}}$ characterizes main image structures. We employ two network branches to deal with structures and details respectively. For the detail branch, we minimize the $L1$ distance between the detail and the output of the high-frequency branch to preserve gradient discontinuities, while for the structure branch we enforce $L2$ loss to encourage global smoothness
\begin{equation}
L_{d} = || f(\mathbf{I})-\tilde{\mathbf{D}}||_{1},
\end{equation}
\begin{equation}
L_{s} = || g(\mathbf{I})-\tilde{\mathbf{S}}||_{2},
\end{equation}
where $f(\cdot)$ and $g(\cdot)$ represent the detail branch and the structure branch, respectively. Furthermore, to ensure the fidelity and structure integrity in the composited derained image, we combine $L_{1}$ loss and SSIM loss to constrain the final result, which can be written as
\begin{equation}
L_{r} = || \mathbf{I}_{p}-\mathbf{I}_{gt}||_{1} + 1-SSIM(\mathbf{I}_{p}, \mathbf{I}_{gt}).
\end{equation}
where $\mathbf{I}_{p}=f(\mathbf{I})+g(\mathbf{I})$ is the output of the whole network as the prediction of the rain-free background.
Given the aforementioned three kinds of loss, the overall loss can be formulated as
\begin{equation}
L_{c} = \lambda_{1}L_{d}+\lambda_{2}L_{s}+\lambda_{3}L_{r},
\end{equation}
where $\lambda_{1}$, $\lambda_{2}$ and $\lambda_{3}$ are the weighting parameters, which in our experiments are all fixed to be 1.

\subsection{Feature-level Frequency Decomposition}
As indicated in \cite{OCT_CONV}, feature maps are also composed of different frequencies. While in low-level vision, as can be observed in Fig. \ref{fig:introduction}, low- and high- frequency feature maps also reflect image structures and details, which are complementary and tightly correlated. To enhance communication across frequencies, we propose a direction-aware Cross-Median Filter (dCMF) to explicitly extract low-frequency components from an entanglement of background features and rain streak patterns with varying falling directions, and an interactive adapter to implicitly enhance feature decomposition through interactive connections.

\textbf{Direction-aware Cross-Median Filter: }Direction-aware Cross-Median Filter (dCMF) aims at separating different frequencies on rain-affected features in the communication paths across the branches. As shown in Fig. \ref{fig:network_framework}, in the HILO module, dCMF extracts low-frequency components, which are adapted by channel-wise attention using Squeeze-and-Excitation  and then sent to the structure learning branch, while in the LIHO module, the residual of the dCMF filtering result is delivered to the high-frequency branch.

\begin{figure}[!t] \centering
	\includegraphics[height=0.8\linewidth, width=1\linewidth]{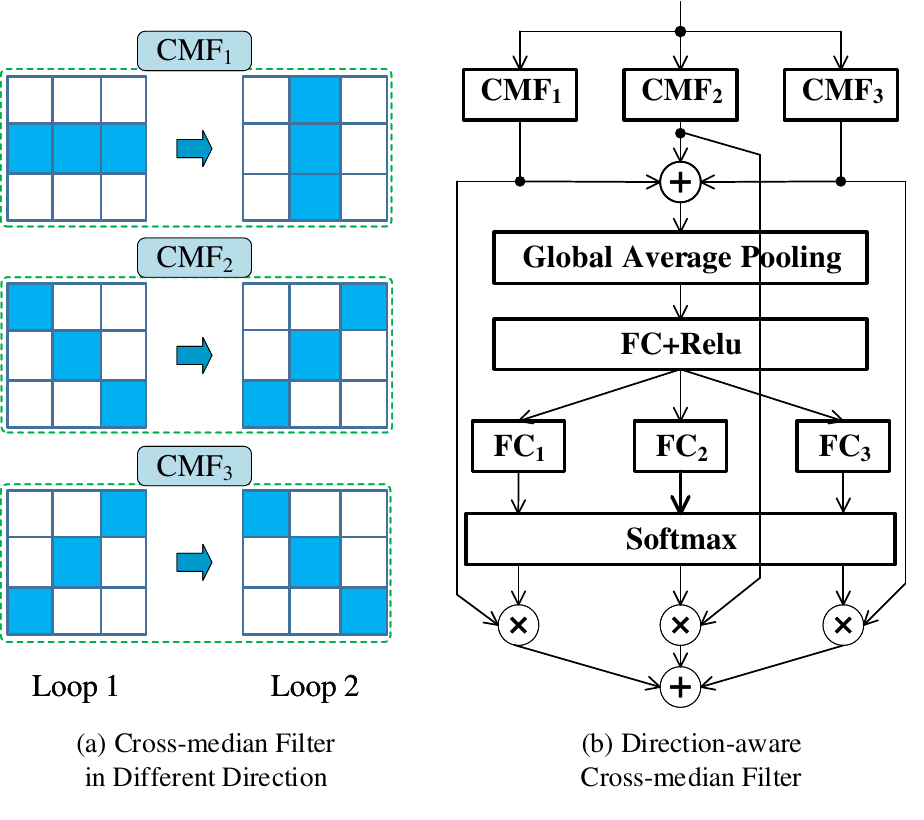}
	\caption{The detailed structure of direction-aware Cross-Median Filter (dCMF), where the self-attention module described in (b) determines the weights of each direction in (a).}
	\label{fig:dcmf}
\end{figure}

Our key observation in designing dCMF is that, rain patterns, not only in rainy images but also in their feature maps, have the properties of relatively high intensity and globally consistent orientation, which makes them distinguishable to background patterns in similar scales (see Fig. \ref{fig:introduction}).
To take full advantage of this prior, we make three additional modifications upon a naive low-pass image filter. First, we adopt median pooling rather than averaging to avoid extreme values brought by rain streaks in feature maps.  Second, we replace 2D filtering kernels with 1D directional lines. It is intuitive that rain mainly falls in the vertical direction and leaves traces of globally vertical streaks. Suppose a 1D filtering line centered at a single rain streak, and the angle between them is denoted as $\alpha$. It is clear that, when $\alpha=90^\circ$ the filtering kernel is least affected by this rain streak and other possible rain directions in this rainy image. However, real scenes are much more complicated, the direction of rain streaks can be affected by many factors, such as wind and obstacles. To take this complexity into consideration, we enumerate different orientations of the 1D kernels as shown in the first column of Fig. \ref{fig:dcmf} (a), extending $CMF_{1}$ to the set $\{CMF_{1},CMF_{2},CMF_{3}\}$ and leverage a self-attention mechanism to learn the importance of each direction. To enforce attention on both feature channels and different groups of filtering results, we adopt a similar strategy with \cite{li2019selective}: three groups of filtering results are aggregated through addition and global average pooling for computing the individual attentions for each group, and the weighted sum of the groups constitute the final output of dCMF. Third, each 1D kernel in $CMF_{i}$ is followed  by  a  crisscross  counterpart to constitute a complete Cross-Median Filter (CMF). Each CMF has exactly the same receptive field  with the corresponding 2D kernel, but is more robust against rain streaks since the second kernel operates on the feature maps with greatly reduced rain patterns. 



\textbf{Interactive Adapters: }For frequency exchange across branches, dCMF leverages prior knowledge on rain streaks to explicitly compute low-frequency components, while the interactive adapter uses learnable convolutional kernels as the frequency filter. The behavior of the interactive adapter is guided by the decomposed labels as a complementary to dCMF-based frequency decomposition.
As shown in Fig. \ref{fig:network_framework}, we adopt asymmetric convolution blocks (ACB) \cite{ACNET} to integrate features from different branches and automatically adjust the information exchange between branches. Suppose $z_{d}, z_{s} \in R^{F \times H \times W}$ are the input features of the adapter, ${\Psi}_{d_{i}}$ and  ${\Psi}_{s_{i}}$denotes the ACB unit, $d$ and $s$ refer to the detail branch and the structure branch, respectively. The basic function of the interactive adapter can be expressed as
\begin{equation}
z_{d}' = S({\Psi}_{d_{1}}(z_{d})) + \Psi_{s_{2}}(z_{s})),
\end{equation} 
where $S(\cdot)$ refers to batch normalization followed by rectified linear unit and $z_{d}'\in R^{F \times H \times W}$ is the output feature. For each interactive adapter, the above function is computed twice, and the second output feature can be obtained by simply replace $z_{d},z_{s}$ with $z_{d}',z_{s}'$. Due to the symmetry in interactive adapters, the corresponding function in the structure branch can be easily inferred.
Through these dual interaction functions, redundant information can be efficiently transferred to another path and thus encourages exploration on new features. In the adapter, computation occurs in parallel at both the dual branches and the interactive paths, which allows accurate decomposition in an information intensive but computation efficient way.

\begin{figure*}[!t] \centering
	\includegraphics[width=0.95\linewidth]{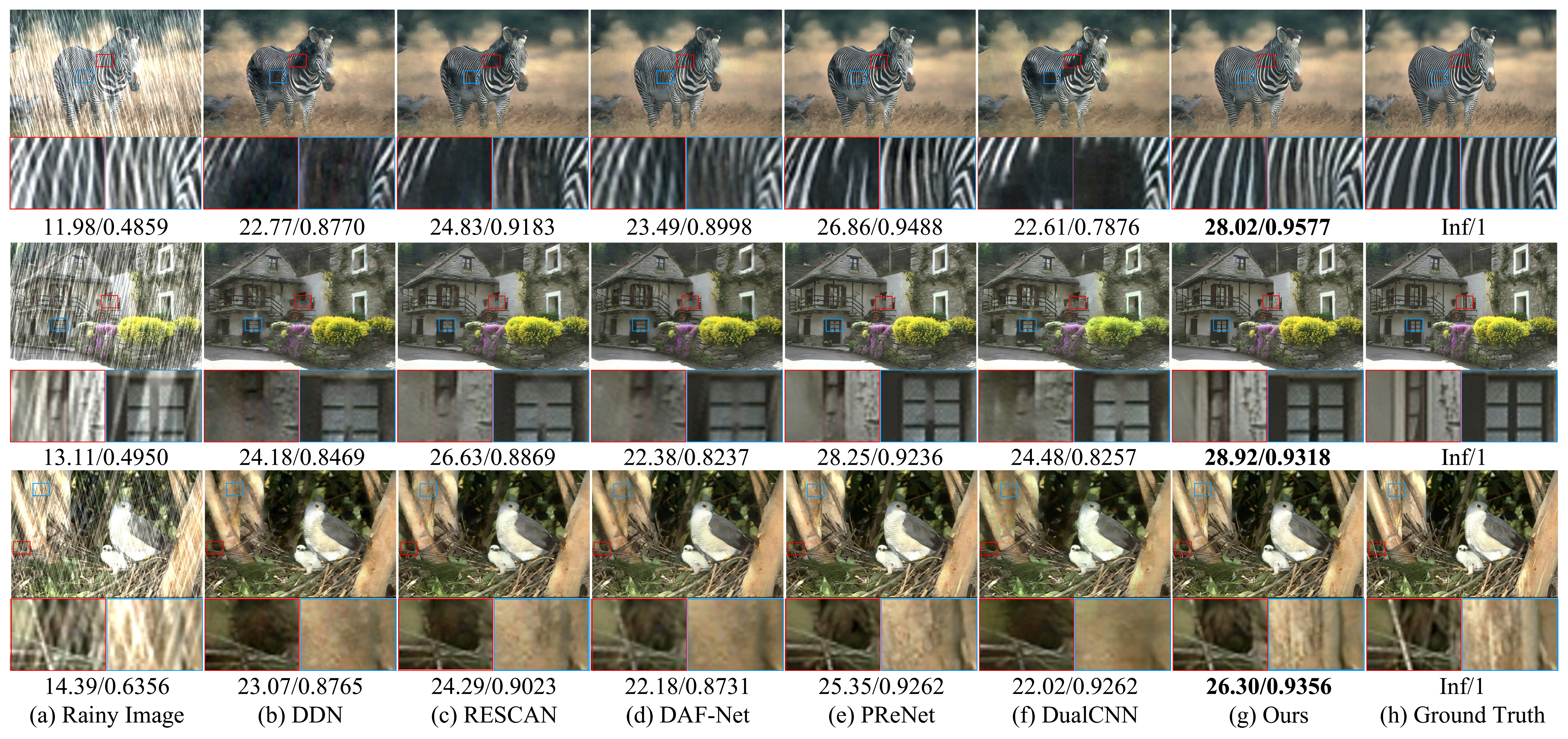}
	\caption{
		Image deraining results tested on the synthetic datasets.  From (a)-(h): (a) the rainy images, and the deraining results of (b) DDN, (c) RESCAN, (d) DAF-Net, (e) PReNet, (f) DualCNN, (g) ours and (h) the ground truth, respectively.}
	
	\label{fig:sys}
\end{figure*}

\begin{figure*}[!t] \centering
	\includegraphics[width=0.95\linewidth]{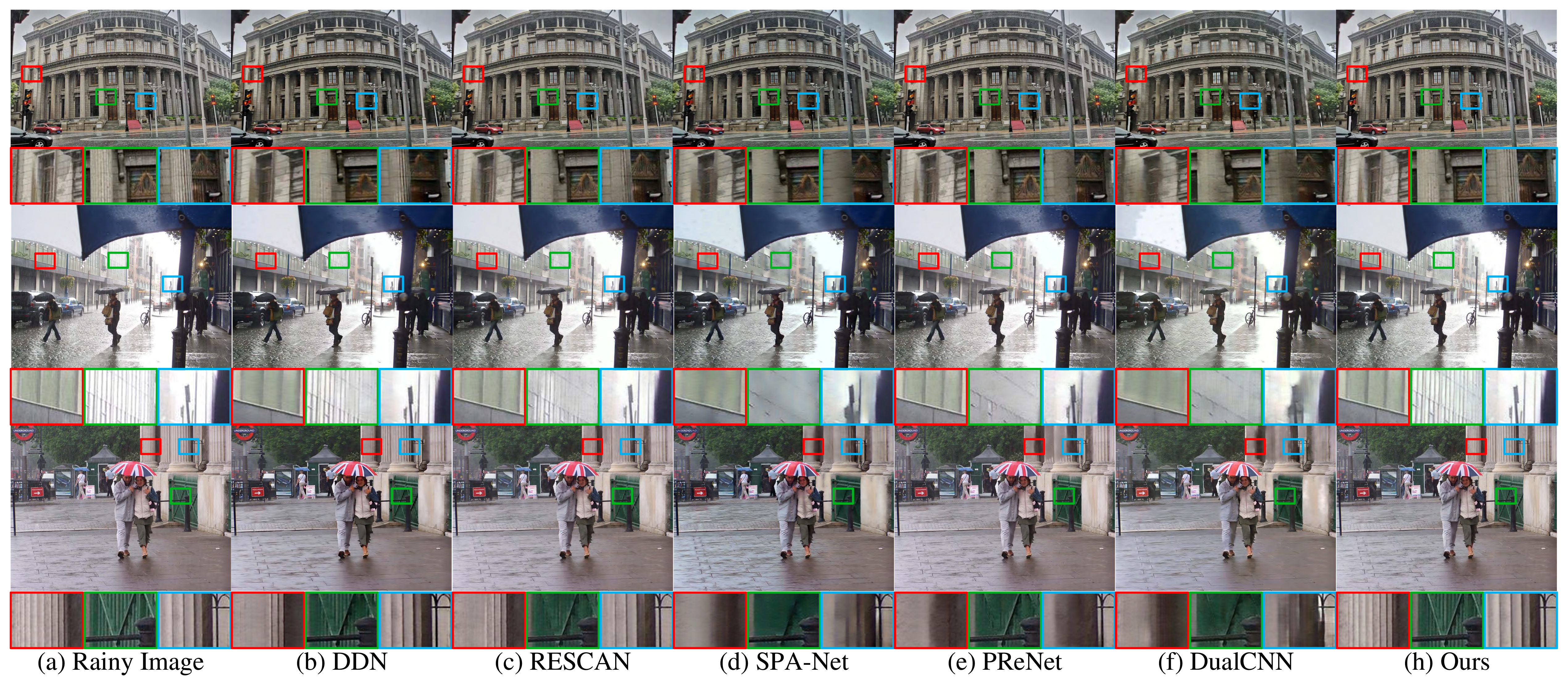}
	\caption{Image deraining results tested on the real-world datasets. From (a)-(h): (a) the rainy images, and the deraining results of (b) DDN, (c) RESCACN, (d) SPA-Net, (e) PReNet, (f) DualCNN, and (g) ours, respectively.
	}
	\label{fig:real}
\end{figure*}

\section{Experiments and Results}
In this section, we evaluate our method on three synthetic datasets: Rain200L, Rain200H \cite{yang2017deep} and Rain800 \cite{RAIN800}. Since no rain-free ground truths for real-world images are provided, we performed a user study on several real-world datasets . Please refer to the supplementary materials for more details.



\begin{table}[htbp]
\small
	\centering
	\begin{threeparttable}
		\centering
		\setlength{\tabcolsep}{0.8mm}{
			\begin{tabular}{ccccccc}
				\toprule
				\multirow{2}{*}{Dataset}&
				\multicolumn{2}{c}{Rain200L}&\multicolumn{2}{c}{Rain200H}&\multicolumn{2}{c}{Rain800}\cr
				\cmidrule(lr){2-3} \cmidrule(lr){4-5} \cmidrule(lr){6-7}
				& PSNR & SSIM & PSNR & SSIM & PSNR & SSIM \cr
				\midrule
				GMM   & 27.16 & 0.8982 & 13.04 & 0.4673 & 24.04 & 0.8675 \cr
				DSC     & 25.68 & 0.8751 & 13.17 & 0.4272 & 20.95 & 0.7530 \cr
				DDN  & 33.01 & 0.9692 & 24.64 & 0.8489 & 24.04 & 0.8675 \cr
				DualCNN  & 32.93 & 0.9575 & 24.09 & 0.7632 & 23.83 & 0.8395 \cr
				RESCAN    & 37.07 & 0.9867 & 26.60 & 0.8974 & 24.09 & 0.8410 \cr
				RWL  & 36.75 & 0.9632 & 26.89 & 0.8406 & 27.79 & 0.8795 \cr
				DAF-Net  & 32.07 & 0.9641 & 24.65 & 0.8607 & 25.27 & 0.8895 \cr
				SPA-Net & 31.59 & 0.9652 & 23.04 & 0.8522 & 22.41 & 0.8382 \cr
				PReNet  & 36.76 & 0.9796 & 28.08 & 0.8871 & \underline{26.61}& 0.9015 \cr
				DRD-Net & \underline{37.15} & \underline{0.9873} & \textbf{28.16} & \underline{0.9201} & 26.32 & \underline{0.9018} \cr
				Ours  & \textbf{37.74} & \textbf{0.9896} & \underline{28.09} & \textbf{0.9316}& \textbf{26.86}& \textbf{0.9164} \cr
				\bottomrule
			\end{tabular}
		}
	\end{threeparttable}
	\caption{Quantitative experiments evaluated on three recognized synthetic datasets. The 1st and 2nd best results boldfaced and underlined.}
	\label{Tab:compare}
\end{table}

		
	

\subsection{Comparison with the State-of-the-Arts}
We compare our method with two traditional methods: GMM \cite{GMM} and DSC \cite{DSC}, and six learning-based methods: DDN \cite{DDN}, RESCAN \cite{RESCAN}, DualCNN \cite{DUAL}, DAF-Net \cite{DAF-Net}, SPA-Net \cite{SPA-Net}, PReNet \cite{PRENET}, and DRD-Net \cite{d_deng2019drd}.

The quantitative evaluation results of PSNR and SSIM are shown in Tab. \ref{Tab:compare}. As can be observed, our proposed method mostly obtains the highest values of PSNR and SSIM than other methods on the synthetic datasets. The visual comparisons are shown in Fig. \ref{fig:sys}, from which one can observe that our method better remains the structure and preserves the detail of images. 


Furthermore, visual evaluation on a series of real-world rainy mages is provided in Fig. \ref{fig:real}, from which one can observe that our method can not only remove real rain streaks but also better preserve the image structures and details. As can be seen, challenging areas, such as the textures of the pillars and the border of the wall, are well preserved by our method.

\subsection{Ablation Study}

\textbf{Ablation Study on Different Components:} In Tab. \ref{Tab:ablation_1}, we show quantitative results in order to validate the effectiveness of: dual branch architecture, interactive adapter, direction-aware Cross-Median Filter in HILO and LIHO modules.

\begin{itemize}
	\item \textbf{BL:} Baseline (BL) indicates that we use a single branch with the residual network to learn a rainy-to-derained function.
	
	\item \textbf{DBL:} Dual Baseline (DBL) indicates that we use two same branches without interaction for single image rain removal, which learns the detail image and the structure image respectively.
	
	
	\item \textbf{DBL+I:} Replacing the residual block with Interactive Adapter in DBL (We remove the HILO and LIHO from our proposed network).
	
	
	\item \textbf{DBL+I+O:} Adding HILO, LIHO and OC to DBL ( Interactive adapter is replaced by the Octave Conv (OC) in our network).

	
\end{itemize}



\begin{table}[htbp]
\small
	\centering

	\begin{threeparttable}
		\centering
		\setlength{\tabcolsep}{0.5mm}{
			\begin{tabular}{cccccccc}
				\toprule
				Dataset & Metrics & BL & DBL & DBL+I & DBL+I+O & Ours \cr
				\toprule
				\multirow{2}{*}{Rain200L}
				& PSNR & 35.57 & 36.33 & 36.96 & \underline{37.33}  & \textbf{37.74}  \cr
				& SSIM & 0.9759 & 0.9864  & 0.9879& \underline{0.9889}  & \textbf{0.9896}  \cr
				\multirow{2}{*}{Rain200H}
				& PSNR & 26.20 & 27.03 & \underline{27.95} & 27.58  & \textbf{28.09}  \cr
				& SSIM &0.8245 & 0.9212 & \underline{0.9310}  & 0.9260 & \textbf{0.9316}  \cr
				\multirow{2}{*}{Rain800}
				& PSNR & 25.16 & 25.23  & 25.46 & \underline{26.19} & \textbf{26.84}  \cr
				& SSIM & 0.9008 & 0.9043  & 0.9034 & \underline{0.9086}  & \textbf{0.9164}  \cr
				\bottomrule
			\end{tabular}
		}
	\end{threeparttable}
	\caption{Quantitative comparison between our network and other network architectures on Rain200H.}
	\label{Tab:ablation_1}
\end{table}

\begin{table}[htbp]
\small
	\centering

	\begin{threeparttable}
		
		\centering
		\setlength{\tabcolsep}{2.8mm}{
			\begin{tabular}{ccccc}
				\toprule
				Datasets&  Metrics& MF& Gaussian & Our \cr
				\toprule
				\multirow{2}{*}{Rain200H}
				& PSNR & 27.86 & 22.58 & 28.09 \cr
				& SSIM & 0.9290 & 0.8064 & 0.9316\cr
				\bottomrule
			\end{tabular}
		}
	\end{threeparttable}
	\caption{Quantitative evaluation. The result of our method by replacing the dCMF with the MF and the Gaussian filter on Rain200H, respectively.}
	\label{Tab:CMIB_with_cmf}
\end{table}

\textbf{Analysis on dCMF: } To validate the effectiveness of our proposed dCMF in HILO and LIHO, we remove them from our method and the result can be found in the 'DBL+I` column of Tab. \ref{Tab:ablation_1}. In addition, we replace dCMF with ordinary $3\times3$ kernel median filter and Gaussian filter, as shown in Tab. \ref{Tab:CMIB_with_cmf}. Generally speaking, both results show clear advantage of dCMF in the deraining task, and we also visually inspected that deraining results without dCMF suffer from heavier degradations.

\textbf{Analysis on Interactive Adapter: } In order to further analyze the necessity of the  interactive adapter, we replace it with another frequency decomposition method: octave convolution (OC) \cite{OCT_CONV}, in our network and the result is shown in Tab. \ref{Tab:ablation_1}, which demonstrates that not only can the interactive blocks between two branches improve the performance of the network but also can our interactive adapter outperforms the Octave Conv in the deraining task.

%

\subsection{Running Time}
We compare the running time of our method with different approaches on Rain200H. As shown in Tab. \ref{Tab:time}, our method is not the fastest one, but reaches a reasonable balance between performance and efficiency.

\begin{table}[htbp]
\small
	\centering
	
	\begin{threeparttable}
		
		\footnotesize
		\centering
		\setlength{\tabcolsep}{0.6mm}{
			\begin{tabular}{ccccccccc}
				\toprule
				Metrics  & DSC & DDN & \tabincell{c}{RES\\CAN} & \tabincell{c}{Dual\\CNN} & \tabincell{c}{DAF\\Net}&  \tabincell{c}{PRe\\Net} & \tabincell{c}{SPA\\Net} & Ours \cr
				\toprule
				PSNR  & 13.17 & 24.64 & 26.60 & 24.09 & 24.65 & 28.08& 23.04& 28.09 \cr
				SSIM  & 0.4272 & 0.8489 & 0.8974 & 0.7632 & 0.8607 & 0.8871& 0.8522& 0.9316 \cr
				
				\tabincell{c}{Avg\\time}   & 92.9s & 0.03s & 0.25s & 0.06s & 0.52s & 0.20s& 0.06s& 0.31s \cr
				\bottomrule
			\end{tabular}
		}
	\end{threeparttable}
	\caption{Averaged time (in seconds) and performances of different methods on Rain200H. }
	\label{Tab:time}
\end{table}



\section{Conclusion}
We propose an interactive dual-branch network where features of different frequencies are learned and exchanged to enhance the performance of single image deraining. The communication between high- and low- frequency branches relys on two key designs: (1) instead of using convolutional filters consistent in all directions, we propose direction-aware Cross-Median Filter to thoroughly purge rain patterns in frequency decomposition; (2) we present the interactive adapter to enhance feature learning and interaction towards decomposed labels.

\section*{Acknowledgements}
This work was supported by the National Natural Science Foundation of
China (Nos. 62032011, 61502137).

\bibliographystyle{named}
\bibliography{bib21}

\end{document}